# Universal Induction with Varying Sets of Combinators

Alexey Potapov[1,2], Sergey Rodionov[1,3]

[1]AIDEUS, Russia
[2]National Research University of Information Technology, Mechanics and Optics, St. Petersburg, Russia
[3]Aix Marseille Université, CNRS, LAM (Laboratoire d'Astrophysique de Marseille) UMR 7326, 13388, Marseille, France
{potapov,rodionov}@aideus.com

Universal induction is a crucial issue in AGI. Its practical applicability can be achieved by the choice of the reference machine or representation of algorithms agreed with the environment. This machine should be updatable for solving subsequent tasks more efficiently. We study this problem on an example of combinatory logic as the very simple Turing-complete reference machine, which enables modifying program representations by introducing different sets of primitive combinators. Genetic programming system is used to search for combinator expressions, which are easily decomposed into sub-expressions being recombined in crossover. Our experiments show that low-complexity induction or prediction tasks can be solved by the developed system (much more efficiently than using brute force); useful combinators can be revealed and included into the representation simplifying more difficult tasks. However, optimal sets of combinators depend on the specific task, so the reference machine should be adaptively chosen in coordination with the search engine.

## 1 Introduction

Universal algorithmic induction or prediction based on Kolmogorov complexity or Solomonoff probability is one of key components of mathematical models of AGI [1]. Of course, direct application of pure universal induction is impractical in general. One should introduce some strong bias and priors to turn it into efficient pragmatic AGI [2]. At the same time, some cases, for which no priors are given, will always be encountered. True AGI should be able to deal with these cases.

Thus, universal induction can still be an essential component of pragmatic AGI. One can even assume that it is the basic building block of intelligence. For example cortex columns might perform universal induction with low-complexity models. Some such block of universal induction should be presented even in complex pragmatically but generally intelligent systems with numerous priors. Unsurprisingly, different authors have introduced search for short programs or codelets both for environment models and behavior policies in their models or cognitive architectures intended not only for theoretical analysis (e.g. [3–5]).

Choice of the reference machine greatly influences efficiency of universal induction [4], [6]. However, if universal induction should work in cases, for which no priors exist, it might be inessential, what reference machine to use in low-complexity "unbiased" induction (especially because one universal machine can be emulated on another one). At the same time, when it comes to choosing the very concrete implementation of the reference machine in practice, it appears that intuitively simple regularities cannot be induced with the use of arbitrary reference machines. Additionally, some special programs such as self-interpreters have considerably different lengths within different formalisms [7]. Analogy with evolution shows that this difference can be crucial. Indeed, emergence of self-replicants enabling further incremental self-improvement is possible, only if they are not too complex. Thus, the choice of the reference machine does matter, even if we consider low-complexity universal induction without concern for pragmatic priors.

In practice, one would like to utilize some advanced programming language as the reference machine to be able to solve at least slightly interesting induction tasks, but the choice of language details is usually very loosely grounded. In theory, some universal Turing machine without its complete definition is frequently used as the reference machine (e.g. [1], [6]). Gradual transition between these two extrema should exist. What general requirements can be put on the reference machine?

Direct search for long algorithms is practically impossible. Fortunately, any algorithm can be represented as a concatenation of its parts. For example, algorithms can be represented as strings encoding programs for universal Turing machine. These strings can easily be divided into substrings, and one can try to find them separately or at least without enumeration of combinations of all possible substrings. However, independent reconstruction of such substrings will be possible, only if each of them has it own utility. Apparently, possibility of decomposition depends on the task being solved, but it also depends on the used reference machine (or representation of algorithms). For example, arbitrary substring of a program for UTM encoded in some binary form may have no self-contained sense.

Some other formalisms such as recursive functions and λ-calculus include composition as the basic operation. Since crossover in genetic programming (GP) allows programs to exchange their subprograms, GP and these formalisms ideally fit each other. Indeed, traditional functional programming languages are built on the basis of λ-calculus and are frequently used in functional GP, particularly in application to inductive programming, in which, however, strong biases (uninteresting for us here) are usually introduced in order to achieve feasible solutions of such tasks as number sequences prediction. Also, application of GP to both functional programming languages and pure λ-calculus requires dealing with local variables that makes it tricky. Another formalism, which is extensionally equal to λ-calculus, and can represent programs that use variables (e.g. via λ-abstractions, let-expressions, etc.) without introducing variables (similar to point-free style), is combinatory logic (CL). This makes implementation of GP with CL rather simple, because it doesn't require using complex genetic operators for handling variables, yet as powerful as with λ-calculus.

Surprisingly, there are few implementations of GP with CL in comparison with λ-calculus. Additionally, CL is typically used with GP for automatic theorem provers [8]; as far as we know, there are no solutions for induction on binary strings using CL. At the same time, some authors claim that "combinator expressions are an ideal representation for functional genetic programming", "because they are simple in structure, but can represent arbitrary functional programs", and that "genetic programming with combinator expressions compares favorably to prior approaches, namely the works of Yu [37], Kirshenbaum [18], Agapitos and Lucas [1], Wong and Leung [35], Koza [20], Langdon [21], and Katayama [17]" (see [9] and references ibid). Thus, there is an interesting niche of implementing universal induction with CL.

In this paper, we implement GP search for CL expressions producing given binary strings. However, our task is not to prove favorability of CL or to develop efficient GP system, which are used here as tools. We are interested in stricter and more detailed comparison and selection of representations of algorithms. To do this, we use different sets of primitive combinators within CL that yields different program representations, and compare their performance on different tasks of inductive inference.

## 2 Combinatory Logic as the Family of Reference Machines

Detailed description of combinatory logic goes beyond the scope of our paper, but at least some notions should be introduced in the context of the topic under consideration. Expressions or terms in CL are either atomic symbols (primitive combinators or additional terminal symbols) or a composition of expressions. Computation in CL is represented as the process of reduction of a combinator expression. This process consists of applications of reduction rules defined for each primitive combinator. For example, if two combinators **S** and **K** are given, and their reduction rules have the following form:

$$\mathbf{K}\,x\,y \rightarrow x \qquad\qquad \mathbf{S}\,x\,y\,z \rightarrow x\,z\,(y\,z)$$

where $x$, $y$, $z$ are arbitrary terms, the expression ($\mathbf{S}\,\mathbf{K}\,\mathbf{K}\,x$) will be reduced as follows

$$\mathbf{S}\,\mathbf{K}\,\mathbf{K}\,x \rightarrow \mathbf{K}\,x\,(\mathbf{K}\,x) \rightarrow x$$

It should be pointed out that application is left associative, i.e. ($\mathbf{S}\ x\ y\ z$) is interpreted as $(((\mathbf{S}\,x)\,y)\,z)$.

Interestingly, these **S** and **K** primitive combinators constitute the complete basis, i.e. for any computable function, there is a combinator expression composed of **S** and **K** (and brackets), which is extensionally equivalent to this function.

For example, one can use combinator expressions with additional 0 and 1 as the terminal symbols to generate arbitrary binary strings. The reduction of **S**10(01) will produce 1(01)(0(01)). This gives us the very simple Turing-complete reference machine for studying universal induction/prediction.

Reduction rules for the primitive combinators are also easy to implement, if combinator expressions are represented as trees. Different possible tree representations can be proposed, e.g. one can place all terminal symbols in leaves of a binary tree with each node standing for the application of its left branch to right branch. We implemented another representation, in which each node contains a terminal symbols, and branches correspond to expressions, which this symbol is applied to. One good feature of CL is that any such tree can be considered as valid (possibly not reducible though). This feature makes implementations of different search techniques for combinator expressions including genetic operators also very simple.

Combinatory logic can be interesting as the basic representation of algorithms for universal induction. However, comparison of efficiency of different sets (libraries) of primitive combinators is even more interesting. The most traditional combinators (in addition to **S**, **K**) in CL are describe with the following and some other reduction rules

**I** $x \rightarrow x$ **B** $x\,y\,z \rightarrow x\,(y\,z)$ **C** $x\,y\,z \rightarrow x\,z\,y$
**W** $x\,y \rightarrow x\,y\,y$ **Y** $x \rightarrow x$ (**Y** $x$)

The choice of the library of primitive combinators greatly influences the length of different algorithms, and can be considered as the choice of the reference machine (from some family of such machines). Extension of the library of combinators during the agent's lifetime will correspond to adaptive versions of universal induction or incremental learning (e.g. [10], [11]).

If the same expression appears several times in solutions of consequent induction tasks, it can be moved to the reference machine as the new primitive combinator. Since the same combinator can appear in one solution more than once, it can be useful to try introducing different combinators as a higher level of search instead of identifying them as repeating expressions in solutions. Moreover, if non-exhaustive search is performed, some expressions can be encountered only once and will not be used to form new combinators on the base of the algorithmic complexity criterion only, but these possible combinators may still be useful as search heuristics. For example, mutations in GP usually replace one symbol by another; certain sequence of mutations, which is hardly probably if intermediate results have small values of the fitness function, can turn into one probable mutation given suitable library of combinators. We don't propose a solution of the problem of library combinators learning, but compare results of solving induction problems depending on the used set of combinators.

However, there are also some "technical details", which can affect the comparison results. Different ways of using results of reduction of CL expressions in induction/prediction exist.

- One can introduce terminal (data) symbols, e.g. 0 and 1, and expect strings of these terms as the result of reduction. This approach is very handful in symbolic induction, in which supplementary terms can stand for built-in functions, numbers, etc. (up to some regular programming language, e.g. [8]). Usage of built-in functions in addition to combinators is also more practical, because it allows introducing desirable inductive bias. Another possibility is to interpret combinator symbols in the reduced expression as data elements (e.g. **S** and **K** can stand for 0 and 1). Such expressions as **S**(**K**(**S**(**K**))) cannot be reduced, and arbitrary binary string can be represented in this form. This approach can give more compact descriptions of some regularities, since the same combinator terms can act both as processing elements and as output. It is also not really unnatural, especially if we consider hierarchical representations, in which models of higher levels should produce models of lower levels. Another interesting aspect of this approach is that sensory data can be directly interpreted as code (but usefulness of this interpretation is questionable). One problem here consists in the fact that inverse strings (e.g. 01010… and 10101…) have different complexity that might seem unnatural.

- Reduced expressions can contain remaining combinator symbols, which cannot be found in data strings. One can consider this situation as inconsistency between generated and real strings. However, one can also simply remove unnecessary symbols (e.g. making 010101 and 01S01S01S equivalent). The second approach is somewhat artificial, but it could be more efficient.
- Type theory can be used to determine functional types of expressions (as it is done in [8]). Types are very useful in the case of symbolic induction, in which terminal symbols can have predefined types, but they can be unnecessary, when output strings are composed of combinators or redundant combinators are removed.
- Reduced expressions are trees. Treating them as plain strings is somewhat wasteful, and one would prefer to perform induction on trees (combinators should not be used as output symbols in this case though). However, such redundancy can be very helpful for some search strategies (but of course this will not decrease complexity of the exhaustive search).
- Combinators are traditionally applied to expressions on the right. However, programs and data are not separated in automatically enumerated expressions, especially when combinators are used in output. At the same time, we want to introduce new combinators as expressions appearing in different induction tasks. This means that these expressions should be constant, but should be applied to some variable data. However, variable parts can appear inside constant parts, which will not be combinators in traditional sense. One could forcefully separate "program code" and data (if data is encoded with symbols, which don't act as combinators). Instead, one can also consider trees with arbitrary (not necessarily the rightmost) free leaves.
- Combinators can be applied during reduction in different order. One can start from innermost or outermost (leftmost) combinators (that will be analogues of calls by value and by name). Difference between them is especially prominent, when infinite strings are generated (and this situation is quite typical for the task of prediction). Reduction of outermost combinators can result in exponential slowdown, since it can duplicate unreduced expressions. Indeed, if *z* is an unreduced expression, then applying **S** first in (**S** *x y z*) will result in necessity to reduce *z* twice. At the same time, this can also avoid reducing inner expressions, which will be ignored, e.g. in the case of such expressions as (**K** *x y*). Reduction of innermost combinators is usually more efficient. Unfortunately, it is inappropriate in the task of prediction, when infinite strings are generated, because in such cases it might never reduce leftmost combinators, so prefix of the resulting string will be unknown, e.g.

  $$\mathbf{I}(\mathbf{Y}(01)) \rightarrow \mathbf{I}(01\mathbf{Y}(01)) \rightarrow \mathbf{I}(0101\mathbf{Y}(01)) \rightarrow \ldots$$

  Reduction of outermost combinators allows generating prefixes of infinite strings. Thus, it is more suitable here. We avoided exponential slowdown by implementing lazy computations.

In our implementation, planarization of reduced trees into strings was used, leftmost (or topmost) combinators were applied first, separation of pure combinator expressions (program code) and data wasn't tested in experiments, and types weren't used. Usage of both combinators and data symbols as output was considered. Possibility of ignoring redundant combinator symbols in output strings was also implemented and checked.

## 3 Genetic Programming with CL

The common structure of genetic algorithms is well known. It includes creation of the initial population usually composed of random candidate solutions and generation of consequent populations with the use of crossover, mutations and selection.

In our implementation of GP with CL, crossover is performed by exchanging randomly chosen subtrees in the parent candidate solutions (combinator expressions). For example, these expressions **S(SSS)S(S(S(K(SK))))** and **S(SI)(S(S(SS)))(S(KK))** can exchange the underlined subexpressions producing **S(SSS)S(SI)** and **S(S(S(K(SK))))(S(S(SS)))(S(KK))**. Two types of mutations were implemented. The first type is deletion of a random subtree. The second type is replacement of a terminal symbol in a randomly selected node of the expression with a random terminal symbol. Elitist selection is used with the fitness function that penalizes both complexity of a candidate solution and its precision. Weights of these two summands can be specified on the base of the information criterion within certain coding schemes. However, our experiments involved only data strings, which assume precise models, thus these weights were used as the parameters of GP controlling average length of candidate solutions. There are also such parameters as the size of population, the tournament size (the number of candidate solutions generated before the selection process), and the number of iterations (generations).

Our implementation can be found at https://github.com/aideus/cl-lab

## 4 Experiments

The general framework for using GP with CL has been considered above. Now, let's describe some specific settings, which were used for our experiments. Strings with simple regularities were used, and shortest combinator expressions producing these strings were sought for. We considered selection of one best model per data string, but similar results can be obtained within Solomonoff prediction framework. Population and tournament sizes were 2000 and 4000 correspondingly (other sizes were also tested, but smaller sizes were suitable for easier problems only). Usually, 500 iterations were enough for convergence of the population. GP system was executed many times for each task in order to estimate probability of achieving optimal solution and diversity of final solutions.

The main set of the terminal symbols included **S**, **K**, 0, 1, and combinator symbols were not removed from output strings (results of expression reduction). Results for this set were compared with results for extended sets additionally including **I**, **C**, **B**, **W** or **Y** combinators. Other settings without 0 and 1 terminal symbols and/or with removal of redundant combinator symbols in output strings were also tested. Let's consider the following cases of increasing complexity.

**Constant strings**

At first, the string 000000000000000000 was considered for the *SK01* set. One might assume that constructing the shortest program for this string is the very simple task. However, corresponding programs in the general-purpose languages can be ten or more bytes long, and blind search for them can be rather difficult. GP with CL stably find the solutions like **S**00(**S**00(000)) or **S**(**S**0)0(00000), which are reduced exactly to this string (without continuation). If one takes even more long string containing, e.g. 28 zeros, such solutions as **S**00(**SS**0(**S**0000)0), **S**(**S**(**S**0))0(**SS**000) and **S**00(**S**00(000000)) are found, which produce the required string with 0, 1 and 2 zeros in continuation correspondingly. Further increase of the length of the required string (e.g. up to 80 zeros) allows GP to find such expressions as **S**(**SSS**)**S**(**S**(**S**(**S**0))), **S**(**SSS**)**S**(**S**(**S**(**K**0))) and **SSS**(**SSS**)(**S**(**K**0)), which produce infinite strings of zeros. Moreover, these expressions were found only in ~27% of runs.

Is this result unsuccessful? Actually, infinite continuations are not necessary for prediction, because we should compare lengths of programs, which produce $x$0 and $x$1. Thus, it is normal if the shortest model doesn't produce any continuation. For example, $K_{CL}$(000000000000000000.0) and $K_{CL}$(000000000000000000.1) can be compared, where $K_{CL}$ is the length of the shortest combinator expression producing the given string, and dot is used only to visually separate the string and its supposed continuation. Here, shortest found solutions are **S**(**SS**0)0(0000) and **S**00(**S**00(000))1. Obviously, the first one is shorter and thus more probable. Thus, results for short strings of zeros are perfectly acceptable. Nevertheless, if the shortest model produces supposed infinite continuation, one can guarantee that algorithmic probability of this continuation will be higher than probability of any other continuation. Thus, it is interesting to examine results for constant strings in other settings.

If one would like to generate such strings as KKKKKKKKKKKKKKKKKK or SSSSSSSSSSSSSSSSSS, corresponding solutions will be **SSS**(**SSS**)(**S**(**KK**)) and **SSS**(**S**(**SS**))**S**. GP requires few tens iterations to find the first solution and only two iterations to find the second solution in all runs. These solutions produce infinite strings of K and S in output. This result is more preferable than that in the case of **S**,**K**,0,1 since it ensures very confident prediction.

Quite interestingly, the pattern **SSS**(**SSS**)(**S**(**K**_)) was also encountered in the previous settings. This expression can help to produce any periodic strings. Discovery and inclusion of such expressions into the combinator library is crucial. This expression is similar to the fixed point combinator, yet it doesn't precisely follow the rule **Y***x*→*x***Y***x*. Complexity of discovering such expressions can be different for different initial libraries. Using the extended set of primitive combinators (*SKICBW*), GP finds the solution **SWW**(**B**0) or its counterpart for the string 000000000000000000 in few tens iterations on each run. This solution produces infinite string of zeros.

**Periodic strings**

Let's consider such string as 010101010101010101010101. GP with CL reliably finds combinator expressions in the *SK01* set producing this string, e.g. **S**(**SS**1)1(**S**0010). These solutions don't produce infinite continuations. Again, the string with more than 80 symbols was necessary to find the solution **SSS**(**SSS**)(**S**(**K**(01))), and it was found in less than 10% of runs. However, its discovery in two different tasks allows some higher-level induction procedure to put this expression directly into the library of combinators.

Again, search for SKSKSKSKSKSKSKSKSKSKSKSKSKSKSKSKSKSKSK model appeared to be much simpler requiring few tens iterations to find

**S**(**SSS**)**S**(**S**(**S**(**K**(**SK**)))) on every run. This solution is almost twice simpler than the solution in the previous case, because the alphabet of terminal symbols is smaller.

The solution **SWW**(**B**(01)) or some its counterparts in the case of the *SKICBW01* set is reliably found in 50–100 iterations. Again, this combinator library appears to be more efficient for this task, and the expression **SWW**(**B**_) is the same as in the previous task (it is not the unique solution though).

Of course, the simplest solution can be found if the fixed point combinator is added to the library. Slightly more interesting is that the solution **Y**(01) is found at least two times faster using the set *SKY01* (requiring only 1–2 iterations) than using the set *SKICBWY01*. The fact that additional combinators can increase search complexity is rather obvious, but it can be forgotten in more complicated cases.

Discovery of precise solutions using the *SK01* set for more complex periodic strings like 0111011101110111011101110111101110111 is still relatively simple using GP, but expressions generating infinitely repeating 0111 are very difficult to find. At the same time, expressions generating some true (yet finite) continuation are frequently found, e.g. **S**(**SS**1)(**S**11)(**S**(**SS**)011) produces the given string with 01110111011 as the continuation.

Solutions with infinite continuations are quickly found using the *SKICBW01* set. However, many alternative solutions in addition to **SWW**(**B**(0111)) are found in more than in 90% cases, e.g. **WSW**(**B**(**W**011)) or **SWW**(**B**(**W**011)), which don't contain 0111 explicitly. Consequently, it can be easier to learn some useful expression for the combinator library on simpler tasks and to use them on more difficult tasks (this effect is known in incremental learning, but it is still interesting as higher-order decomposition requiring more detailed study). Again, the solution **Y**(0111) is found in several iterations, if the fixed point combinator is already added into the library.

However, results are not so obvious, if one further increases complexity of strings to be generated. Consider 010110011101000110011101001011001110100011001. Supposed continuation isn't immediately obvious here even for humans. Actually, this string consists of two substrings – (random) 01011001110100011001110101 and its truncated version 0101100111010001100 1. Thus, the most probable continuation should be 11010.

Precise solutions, e.g. **S**00(**S**10(011001(**S**101)0)), are always found using the *SK01* set, and they usually predict the supposed continuation 11010. In rare cases the continuation is 110100. Presence of one additional period in the string to be generated yields similar results, e.g. **S**0(S00)(**S**10(01(10(01(1101))0))). These results are valid, but infinite continuations are very difficult to obtain.

More interestingly, usage of the extended set *SKICBW01* gives no benefits here. Instead, precise solutions, e.g. **W**0(**W**1(**W**(**S**(0110)(**S**(11)01))0)), with adequate continuation are obtained only in approximately 60% runs. These solutions have smaller complexities, so this is the search problem. Even more interestingly, the *SKY01* set yields worse results, which consist in rapid degeneration of the population consisting of imprecise yet very simple solutions like **Y**(**S**0011). The fixed point combinator acts here as "anti-heuristic" for GP.

Thus, it should be concluded that the combinator library cannot be chosen a priori. It should depend both on the task and search strategy.

**More complex regularities**

Discovery of repeating patterns can be achieved by much more efficient means. Algorithmic induction is interesting, because of its potential ability to find arbitrary regularities. Even simplest off-design regularities cannot be discovered by conventional weak machine learning methods. Of course, universal induction is also restricted in discoverable regularities simply because of limited computational resources in practice. However, this limitation can at least be surpassed in limit, while weak methods remain weak independent of the available resources. Thus, it is interesting to consider strings with another underlying regularity

Let's take the string 010010001000010000010000001. Precise solutions, e.g. S0(**SS**(**S**(**S**0)0)(**SS**1000))1, are sometimes found (less than in 20% runs), but they don't make precise predictions. For example, this solution produces 0001 in continuation. Of course, as it was pointed out above, the simplest solution shouldn't necessarily have precise continuation in order to make correct predictions on the base of algorithmic probability. Nevertheless, this result is far from ideal. Although it is expected since infinite continuations are difficult to achieve in these settings for simpler cases.

At the same time, such ideal solutions as **S**(**S**(**SKK**))(**S**(**S**(**SS**)))(**S**(**KK**)) for the string SKSKKSKKKSKKKKSKKKKKSKKKKKK are surprisingly easy to find (its equivalents are found in more than 75% of GP runs). This solution generates infinite precise continuation SKKKKKKKS… The equivalent result **SWBS**(**WS**(**BK**)) can be obtained also using the *SKIBCW* set with similar easiness. Somewhat more difficult yet quite probably (~50%) is to find the solution like **W**(**SIWS**(**B**1))(**WS**(**B**0)) that produces precise infinite binary string in the case of the *SKICBW01* set. Additional inclusion of **Y** combinator makes results worse in this case.

Search for precise solutions using *SK*01 appeared to be unsuccessful for tasks of higher complexity. The string 0100110001110000111100000111110000001111111 was tested, and only imprecise solutions such as 0(**S**11(**S**(**S**0)0(**S**0001111))) producing, e.g. 010000001111100001111100000111111000000111100.001111000001111 (with 5 mistakes) were found. The only precise result **I**(**WI**(**S**(**S**(**CB**)**B**)(**B**(**SB**))))**K** was obtained using *SKICBW* and additionally ignoring redundant combinators (except **S** and **K**) in output producing SKSSKKSSSKKK… with infinite continuation. This and more complex tasks can probably be solved using some additional combinators (or more advanced search strategy).

It should be pointed out that some tasks successfully solved by GP weren't solved by the exhaustive search in reasonable time, while no opposite cases were encountered. Interestingly, effects of extension of the combinator library on exhaustive search performance weren't so unambiguous. However, detailed analysis of the search problem goes beyond the scope of the paper.

## Conclusion

Combinatory logic was considered as the reference machine adjustable by selection of the set of primitive combinators. The genetic programming system was implemented for searching the shortest combinator expressions generating required binary strings.

Performance of GP with CL was evaluated for minimal and extended sets of combinators on different induction tasks. Experiments showed that extension of the combinator library allows GP to solve more difficult induction tasks. At the same time, such extension reduces performance on some tasks. This reduction is usually much less significant than benefits on other tasks, but choosing the best set of combinators as an upper level of search can be useful, especially if the basic search for combinator expressions is non-exhaustive.

Since CL with different sets of primitive combinators can be treated as different reference machines, one can perform extension of combinator library to implement adaptive universal induction. Automation of this process is the main topic for further investigations.

## Acknowledgements

This work was supported by the Russian Federation President's grant Council (MD-1072.2013.9) and the Ministry of Education and Science of the Russian Federation.